\pdfoutput=1
\documentclass[11pt]{article}
\usepackage{acl}
\usepackage{times}
\usepackage{latexsym}
\usepackage[T1]{fontenc}
\usepackage[utf8]{inputenc}
\usepackage{microtype}
\usepackage{multicol}
\usepackage{multirow}
\usepackage{graphicx}
\usepackage{booktabs}

\DeclareUnicodeCharacter{01B0}{{\`{\^o}}}
\DeclareUnicodeCharacter{01A1}{{\~{\^e}}}
\DeclareUnicodeCharacter{0300}{`}

\definecolor{galician}{HTML}{0066CC}
\definecolor{spanish}{HTML}{CC0000}

\title{Cross-lingual Inflection as a Data Augmentation Method for Parsing}
\author
{
Alberto Muñoz-Ortiz, Carlos Gómez-Rodríguez and David Vilares\\
Universidade da Coru\~{n}a, CITIC \\
Departamento de Ciencias de la Computación y Tecnologías de la Información \\
Campus de Elvi\~{n}a s/n, 15071 \\ A Coru\~{n}a, Spain \\
\texttt{\{alberto.munoz.ortiz, carlos.gomez, david.vilares\}@udc.es} \\
}

\begin{document}
\maketitle
\begin{abstract}
We propose a morphology-based method for low-resource (LR) dependency parsing. We train a morphological inflector for target LR languages, and apply it to related rich-resource (RR) treebanks to create cross-lingual (x-inflected) treebanks that resemble the target LR language. We use such inflected treebanks to train parsers in zero- (training on x-inflected treebanks) and few-shot (training on x-inflected and target language treebanks) setups. The results show that the method sometimes improves the baselines, but not consistently.
\end{abstract}

\section{Introduction}

Dependency parsers \cite{dozat-etal-2017-stanfords,ma-etal-2018-stack,strzyz-etal-2019-viable} already achieve accurate results for certain setups \cite{berzak-etal-2016-anchoring}. Yet, they require large amounts of data to work, which hurts low-resource (LR) scenarios. In this line, authors have studied how to overcome this problem.

On data augmentation, recent approaches have replaced subtrees of sentences to generate new ones  \cite{vania-etal-2019-systematic,dehouck-gomez-rodriguez-2020-data}. On cross-lingual learning, authors have explored delexicalized approaches from rich-resource (RR) treebanks. \cite{mcdonald-etal-2011-multi,falenska-cetinoglu-2017-lexicalized}. \citet{wang-eisner-2018-synthetic} permuted constituents of distant treebanks to generate synthetic ones that resembled the target language. \citet{vilares-etal-2016-one,ammar-etal-2016-many} merged treebanks to train multilingual parsers that sometimes could outperform the equivalent monolingual version, which has applications for less-resourced parsing. In the context of multilingual representations, \citet{mulcaire-etal-2019-low} trained a zero-shot parser on top of a polyglot language model, relying on merged RR treebanks too. 

\begin{figure}
    \centering
    \includegraphics[width=0.35\textwidth]{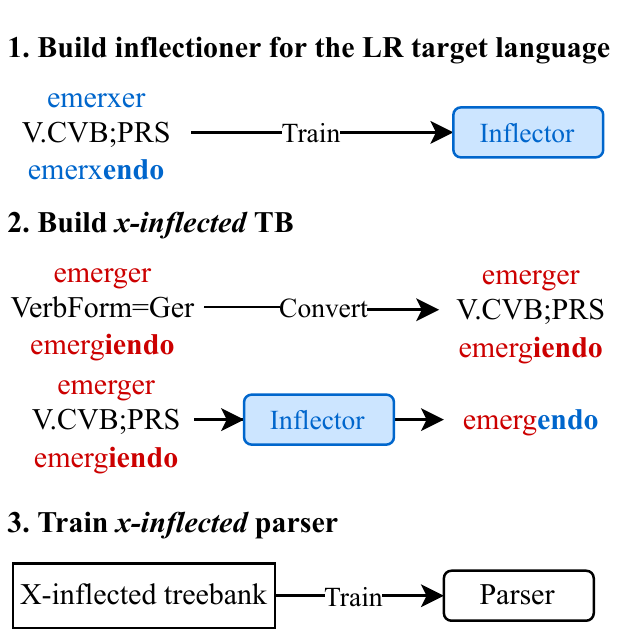}
\caption{X-inflection process for a target LR language (\textcolor{galician}{Galician}) using a source RR treebank (\textcolor{spanish}{Spanish}).}
    \label{fig:morph_process}
\end{figure}

In other matters, morphological inflection \cite{cotterell2016sigmorphon,pimentel2021sigmorphon} generates words from lemmas and morphological feats (e.g. look $\rightarrow$ looking). Also, it is known that morphology helps parsing and that morphological complexity relates to the magnitude of the improvements \cite{dehouck-denis-2018-framework}. Yet, as far as we know, there is no work on cross-lingual morphological inflection as a data augmentation method for parsing. Here, we propose a technique that lies in the intersection between data augmentation, cross-lingual learning, and morphological inflection.

\paragraph{Contribution} We introduce a method that uses cross-lingual morphological inflection to generate `synthetic creole' treebanks, which we call \emph{x-inflected} treebanks. To do so, we require a \emph{source} language treebank from a closely-related language (for which lemmas and morphological feats are available), and a morphological inflection system trained for the \emph{target} language. This way, we expect to generate x-inflected treebanks that should resemble \emph{to a certain extent} the target language (see Figure \ref{fig:morph_process}). %Thus, 
The goal is to improve the parser's performance for languages for which little or no annotated data are available, but for which we can train an accurate morphological inflection system that can be later applied to a related RR treebank and resemble the target language. The code is available at \url{https://github.com/amunozo/x-inflection}.

\section{Preliminaries}
We now describe the basics of our work:

\paragraph{Datasets} 
We use UniMorph \citep[UM;][]{mccarthy-etal-2020-unimorph} for morphology and Universal Dependencies \citep[UD;][]{ud2.7}
\paragraph{Key concepts} We call
\textit{inflector} a morphological system that produces a word form from an input lemma and a set of morphological feats in a given language. We call \textit{target UD treebank} each of the LR treebanks where we test our approach. We call \textit{source UD treebanks} the RR treebanks related to a target LR treebank, used to create a cross-lingual inflected treebank, aka \textit{x-inflected treebank}, which results from applying an inflector over the lemmas and feats of a source UD treebank.

\section{X-inflection as data augmentation}
Character-level models, such as the ones used for morphological inflection, identify shared morphemes across languages with overlapping alphabets \cite{lee-etal-2017-fully,vania2020thesis}. Thus, if two languages share a significant amount of lemmas, n-grams or inflections, an inflector for the first language could maybe produce noisy-but-useful inflected forms for lemmas and feats available for the second language. We hypothesize that this idea can be used for syntactic data augmentation in LR scenarios. Under the assumption that an inflector is available for our target LR language (easier than annotating syntactic data), we could use it to transform a related RR treebank, obtaining silver syntactic data that, despite lexical and grammatical imperfections, could help boost performance.

Our method consists of three steps: (i, \S\ref{ssec:train_morph})  training an inflector for a given target language using UM data,  (ii, \S\ref{ssec:transforming}) \emph{x-inflecting} the source UD treebank, cross-lingually applying the inflector trained in (i), and (iii, \S \ref{ssec:training}) training the \emph{x-inflected} parsers. We summarized the process in Figure \ref{fig:morph_process}. 

\subsection{Building the inflectors}
\label{ssec:train_morph}

We train the inflectors using the \citet{wu-etal-2018-hard} model, and leave all the hyperparameters at their default value. It is a seq2seq model that uses a hard monotonic attention mechanism to identify what parts of the input the model should focus on to generate the correct output string. It offers a good trade-off between speed and accuracy, compared with other alternatives that we tested in early experiments \cite{wu-etal-2021-applying}. We train the models on UM data, and for each language, we shuffle and split it 80-10-10 for the training, development and test sets (so lemmas are distributed)\footnote{For languages containing files for different dialects (e.g. Livvi), we concatenated all the forms prior to splitting.}.

\subsection{Building the \emph{x-inflected} treebanks}
\label{ssec:transforming}

This step requires to: (i) transform the feats column of the source UD treebank into a readable format by the inflector (i.e., UM format), to then (ii) apply the inflector to generate the x-inflected word forms, and (iii) format the output into an x-inflected treebank (i.e., going back to the UD format).

\paragraph{Transform UD feats into UM feats} To x-inflect the source treebank, we first need to convert the morphological feats of the UD treebanks to the UM schema, using the converter by \citet{mccarthy2018udw}.

In early experiments, we also trained inflectors directly on UD feats (following \S \ref{ssec:train_morph}), but the results showed that x-inflected parsers trained this way performed worse, so we discarded it.

More specifically, the selected converter creates a mapping between both schemata. Yet, annotation errors and missing values in both schemata, together with disagreements between them, makes the process non-trivial. To counteract this, the approach introduces a language-dependent post-editing process, which consists in an iterative process that analyzes those forms and lemmas present both in UD and UM, comparing their annotations, and creating rules to refine the mappings between schemata. However, this extra refinement process is only available for some languages.

\paragraph{X-inflecting treebanks} 
The lemmas and UM-transformed feats of the source UD treebank are sent to the target LR language inflector.
The x-inflection is not applied to all elements, only to those lemmas of the source UD treebank whose part-of-speech is contained in the UM data of our target language (e.g., verbs or nouns, see details in Appendix \ref{app:um_data}). Then, these x-inflected forms replace the original forms in the source UD treebank.

\subsection{Training the \emph{x-inflected} parsers} 
\label{ssec:training}
We train the parsers with a graph-based (GB) model \cite{dozat-etal-2017-stanfords}. It contextualizes words with bidirectional LSTMs \citep{hochreiter1997long} and computes head and dependent representations for each word. Then, a biaffine transformation of such vectors is used to find the highest scoring parse tree. We also study a sequence labeling (SL) parser \cite{strzyz-etal-2019-viable} as a lower bound. This parser can be seen as a vanilla biLSTM that only needs softmaxes (instead of a biaffine attention module) to predict syntactic labels, using 2-planar encodings \cite{strzyz-etal-2020-bracketing}, that are naturally decoded into to a dependency tree and work more robustly on low-resource setups \cite{munoz-ortiz-etal-2021-linearizations}. 
\section{Experiments}
We test both (i, \S \ref{section-experiment1-zero-shot}) zero-shot and (ii, \S \ref{section-experiment2-few-shot}) few-shot setups. For evaluation, we use unlabeled (UAS) and labeled attachment scores (LAS). Appendix \ref{app:hardware} reports the hardware and costs.

\paragraph{Data}
We use 10 LR and 21 RR treebanks. Although our method can be applied to any pair of treebanks, the availability of UM and UD resources (in the sense of having LR languages in UM and related RR languages in UD) restricts our empirical analysis to Indo-European and Uralic languages (see Table \ref{tab:few-shot_data}). Yet, we have reasonable diversity and degrees of morphological inflection. For our empirical analysis, we use a relaxed definition of the concept LR for Czech and Latin (as the treebanks used are LR but there are RR treebanks for them in UD), and of the concept RR for Scottish Gaelic (as the treebank used is larger than the Welsh one but not RR). See Appendix \ref{app:tb_information} for the details. 

\begin{table}[]
    \centering
    \scriptsize
    \begin{tabular}{lllll}
    \hline
        Group & LR & ISO & RR\\
    \hline
        Iberian & Galician & glg & Spanish, Catalan, Portuguese\\
        N. Germanic & Faroese & fao & Norwegian (nb), Norwegian (nn),\\
        & & & Swedish, Icelandic, Danish \\
        Finno-Ugric & Hungarian & hun & Finnish, Estonian \\
        West Slavic & Czech & ces &Polish, Slovak \\
        South Slavic & Slovenian & slv & Bulgarian, Croatian, Serbian \\
        Romance & Latin & lat & Spanish, Romanian, French, Catalan, Italian\\
        Baltic & Lithuanian & lit & Latvian \\
        Celtic & Welsh & cym & Irish, Scottish Gaelic\\
        Finnic & Livvi & olo & Finnish, Estonian \\
        Finno-Permic & North Sami & sme & Finnish, Estonian \\ \hline
        
    \end{tabular}
    \caption{LR and RR languages used in our experiments. Some LR treebanks come from RR languages (Czech, Latin) to have more samples.
    }
    \label{tab:few-shot_data}
\end{table}

\subsection{Experiment 1: Zero-shot setup}
\label{section-experiment1-zero-shot}
We test if our method improves parsing accuracy under the assumption that there is no available training data in the target language, but there is an UD treebank for a related language, and enough UM data to train an inflector for the target language. Although the selected LR treebanks have a training set, we here do not use them, but we will in \S \ref{section-experiment2-few-shot}.

\paragraph{Setup} 
For each target LR treebank, we first pair them with related source UD treebanks (from 1 to 5)\footnote{Depending on the resource availability in UM and UD.}, such that they all belong to the most restricted phylogenetic group for which UD data is available. We then train our x-inflected parsers and evaluate them on the corresponding target LR treebank. We compare the results against a baseline consisting in models trained on the source RR treebanks.

\paragraph{Results}

\begin{figure}
    \centering
    \includegraphics[width=0.40\textwidth]{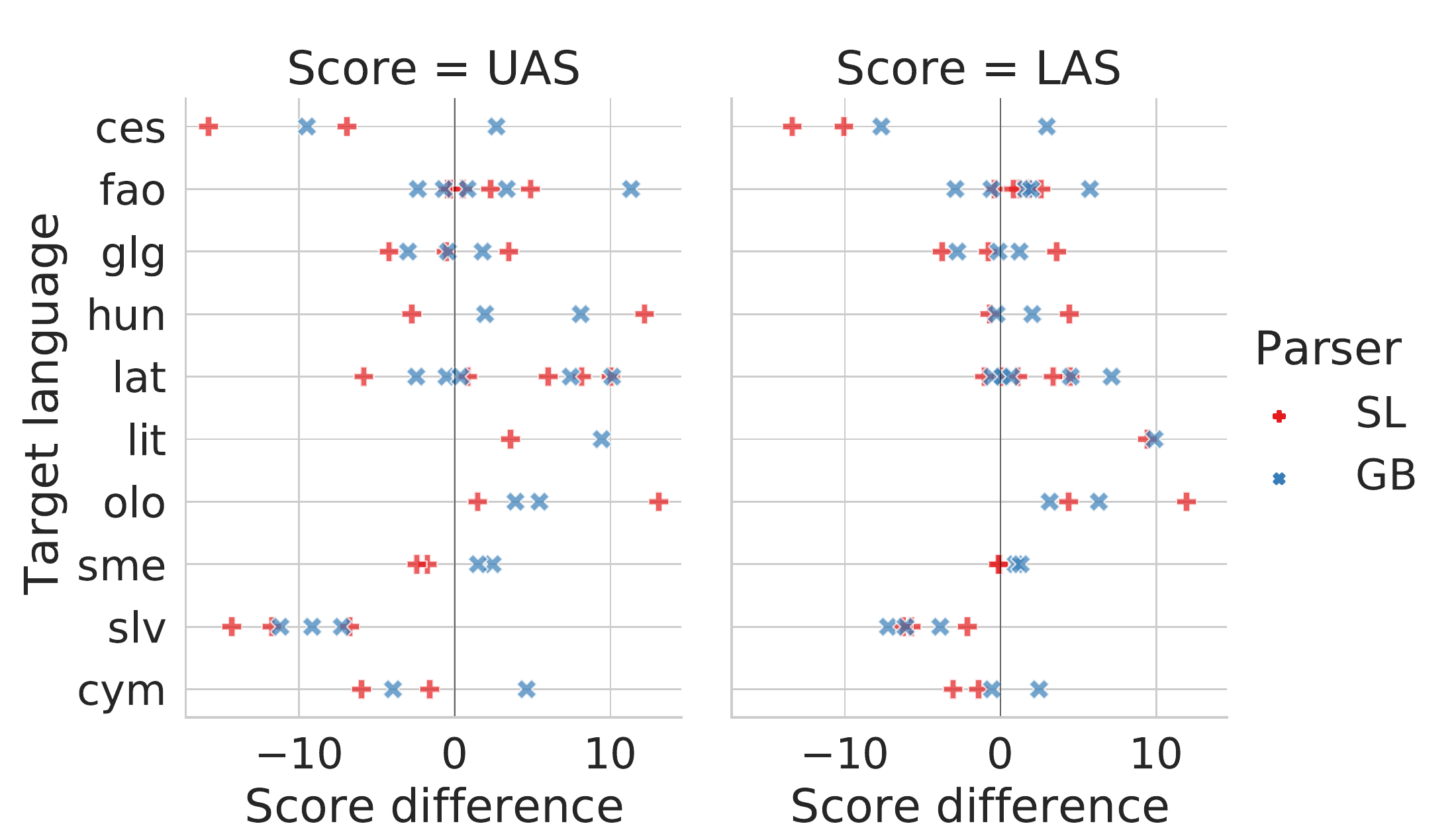}
    \caption{Score differences between x-inflected versions using different source treebanks and the baseline, for both sequence labeling (\textcolor{spanish}{+}) and graph-based (\textcolor{galician}{\texttt{x}}) parsers.}
    \label{fig:zeroshot-scat}
\end{figure}

Figure \ref{fig:zeroshot-scat} shows the results for the zero-shot setup (full results in Appendix \ref{app:exp1_results}). The differences in performance are inconsistent:  for some target LR treebanks the x-inflected models always obtain improvements, e.g. Livvi, for some others the models only obtain decreases, e.g. Slovenian, and for some others there is a mix, e.g. Faroese.

\begin{figure}
    \centering
    \includegraphics[width=0.35\textwidth]{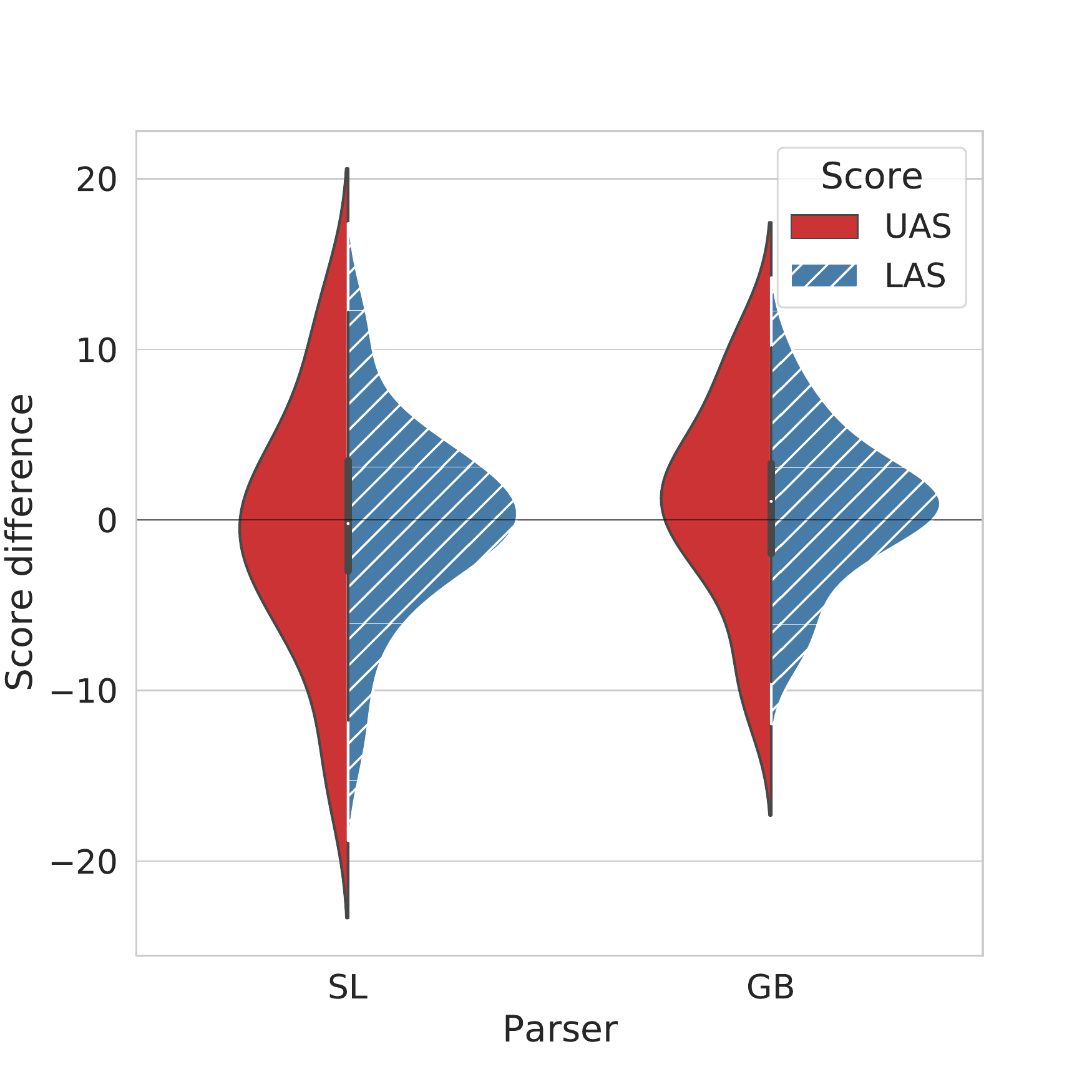}
    \caption{$\Delta$UAS/LAS between the x-inflected models and the baseline for the sequence labeling (SL) and the graph-based (GB) parser.}
    \label{fig:zeroshot-violin}
\end{figure}

Figure \ref{fig:zeroshot-violin} shows the distributions of LAS and UAS differences ($\Delta$) against the baseline versions. For the SL parser, the distribution is centered in 0, with the occurrence of some extreme results. For the GB parser, we see less extreme results and a distribution centered slightly above 0. This suggests that our method could be more effective for the GB approach, but we do not have clear evidence. 

To shed light on what factors might affect the results, Table \ref{tab:exp1_corr} shows the Pearson correlation coefficient (PCC) of the LAS and UAS differences between the x-inflected models and the baselines; \emph{with respect to} features such as the number of forms and lemmas seen in UM training data, feature and lemma overlap between the target and source UD treebanks, or the number of UD training sentences. Although small, the results show some correlations e.g. for the number of forms and lemmas of the UM data ($0.3-0.5$).

\begin{table}[hbtp!]
    \centering
    \scriptsize
    \begin{tabular}{l|rrrr}
\hline
 & \multicolumn{2}{c}{GB}      & \multicolumn{2}{c}{SL}       \\
 & UAS &  LAS &  UAS &  LAS \\
\hline
      \# UM target language forms &  0.31 &     0.34 &  \textbf{0.49} &   \textbf{0.47}\\
      \# UM target language lemmas & 0.32 &     0.32 &   \textbf{0.50} &   \textbf{0.42} \\
      \# UD source treebank training sents. &  0.30 &      0.17 &     0.27 &     0.22\\
      \% Morph. feats shared between treebanks & -0.24 &  -0.24 &    -0.14 &    -0.06\\
      \% Lemmas shared between treebanks & -0.35 &     -0.32 &     -0.20 &    -0.18\\
\hline
\end{tabular}
    \caption{PCC of $\Delta$LAS/UAS between the x-inflected models and the baselines vs different dimensions. Bold numbers represents p-values $<0.05$.}
    \label{tab:exp1_corr}
\end{table}

\subsection{Experiment 2: Few-shot setup}
\label{section-experiment2-few-shot}
Experiment 1 did not show consistent improvements. However, we question whether the reason for our x-inflected models not  consistently improving over the baseline could be that having some annotated data in the target language would help better guide the learning process, or that we are simply not taking advantage of x-inflecting more than one language treebank.
In this line, previous studies have shown that training on harmonized treebanks, i.e. treebanks with the same annotation guidelines but coming from different languages, could improve performance over the corresponding monolingual model \cite{vilares-etal-2016-one}, which has applications to less-resourced languages \cite{ammar-etal-2016-many}.

\paragraph{Setup} 
To test this, we train models on many x-inflected treebanks and evaluate them on the corresponding target LR test sets. Here, we also consider merging the available training data for the target LR language, to have a better understanding of how our approach behaves in few-shot setups. Particularly, we combine all the x-inflected treebanks from the phylogenetic groups described for the previous experiment (see again Table \ref{tab:few-shot_data}), instead of training separate models for each one. We compare the performance against two baselines: (i) models trained on the target LR language training set, and (ii) models trained on a merged training treebank composed of the training set of the target LR treebank and the original training sets of the source treebanks of the related languages (but without x-inflecting them).

\paragraph{Results}
\begin{figure}
    \centering
    \includegraphics[width=0.48\textwidth]{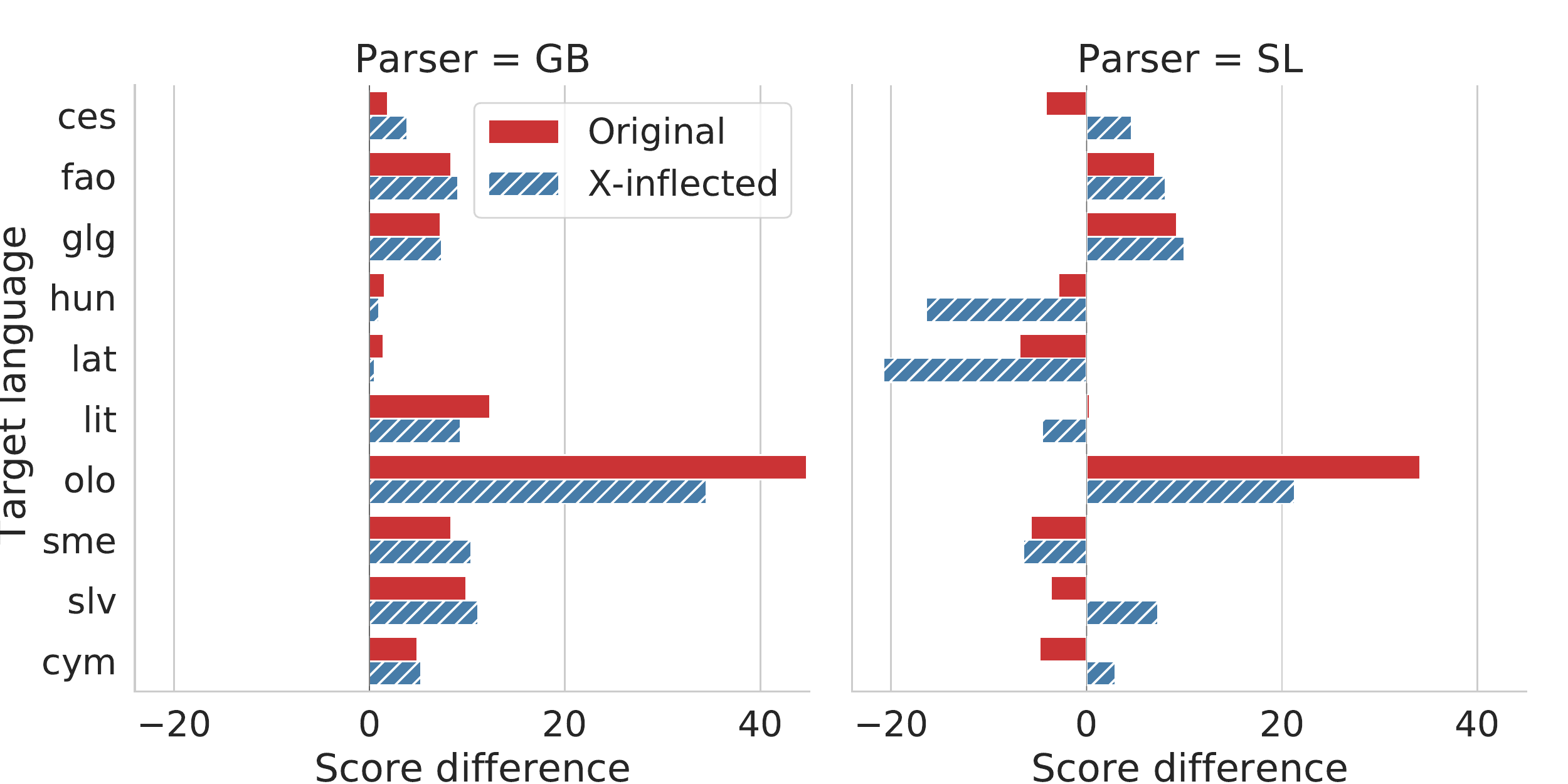}
    \caption{$\Delta$LAS %difference 
    between the models trained on the original and x-inflected groups \emph{with respect to} the model trained on the LR treebank, for both parsers. UAS shows the same tendency as seen in Appendix \ref{app:exp2_results}.}
    \label{fig:fewshot_las}
\end{figure}

Figure \ref{fig:fewshot_las} shows the LAS differences between the merged original and x-inflected models \emph{with respect to the} models that are only trained on data coming from the target LR language (UAS results in Appendix \ref{app:exp2_results}). For the GB parser, all models trained on merged (original or x-inflected) treebanks perform better than their counterparts trained only on the LR treebank, suggesting that adding data from similar languages helps the parsers. However, merging non-x-inflected treebanks sometimes outperforms
merging x-inflected treebanks (e.g. Livvi, Lithuanian, and Latin). For the SL parser merging treebanks is not always beneficial compared to training only on the LR training set. We see that the models trained on harmonized (original or x-inflected) treebanks improve only half of the times. Yet, we see some interesting patterns. For instance, when the x-inflection benefits a sequence-labeling model, it also benefits the graph-based one for the same merged treebank, and \emph{vice versa}. Overall, merging x-inflected treebanks is the best option for 6 out of 10 models, although in many cases the differences are small.

\section{Discussion}
The results show that the proposed method is able to improve parsing results for some treebank pairs under both zero- and few-shot setups, but it also obtains decreases for other pairs. Due to the high number of factors involved, we were unable to clearly isolate those that are beneficial and those that are harmful. However, we identified some reasons that could partially explain the behaviour of the method:

\begin{itemize}
\item PCCs from Table \ref{tab:exp1_corr} show that having more UM data is beneficial to obtain better parsing performance, so better inflectors create better x-inflected treebanks.

\item Conversion between UM and UD schemata is non-trivial and dependent on the language pair (see \citet{mccarthy2018udw} for a detailed analysis), and thus incorrect feature conversions could express different morphological information and mislead the parser.

\item Although both UD and UM aim to follow a universal annotation schema, not all languages are annotated exactly in the same way, expressing similar morphological phenomena with slightly different features or omitting some of them. Therefore, even when the conversion between schemata is correct, the annotation discrepancies between languages may confuse the inflector, which again, would output a word whose form would express different morphological information than the original form.
\end{itemize}

\section{Conclusion}
By cross-inflecting a rich-resource UD treebank using an inflector from a low-resource related language, we can obtain silver, syntactically annotated data to train dependency parsers. Although containing noise and grammatical imperfections, we aimed to test whether the approach could boost performance. The results show that it is possible to obtain improvements (but also decreases) both for zero- and few-shot setups.

About this, we could not clearly identify what aspects make the approach succeed or fail. Although we identified moderate correlations between scores and the amount of available UM data for the target language, we hypothesize that other aspects that are hard to measure could be playing a role: (i) incorrect/incomplete feature conversion from UM to UD schemata that might make the cross-lingual inflections carry different information that the inflections in the original language, or (ii) unknown input features for a given inflector due to differences in exhaustiveness between the UM and UD annotations.

\section*{Acknowledgements}
This work is supported by a 2020 Leonardo Grant for Researchers and Cultural Creators from the FBBVA,\footnote{FBBVA accepts no responsibility for the opinions, statements and contents included in the project and/or the results thereof, which are entirely the responsibility of the authors.} as well as by the European Research Council (ERC), under the European Union’s Horizon 2020 research and innovation programme (FASTPARSE, grant agreement No 714150). The work is also supported by ERDF/MICINN-AEI (SCANNER-UDC, PID2020-113230RB-C21), by Xunta de Galicia (ED431C 2020/11), and by Centro de Investigación de Galicia ‘‘CITIC’’ which is funded by Xunta de Galicia, Spain and the European Union (ERDF - Galicia 2014–2020 Program), by grant ED431G 2019/01.

\bibliography{anthology,custom,ranlp2021}
\bibliographystyle{acl_natbib}

\clearpage

\appendix
\section{UM data information}\label{app:um_data}
\begin{table}[hbtp]
    \centering
    \scriptsize
    \begin{tabular}{llcc}
        \hline
        Language & PoS & Lemmas & Forms \\
        \hline
        Galician & V & 486 & 36\,801\\
        Livvi & N, V, A & 23\,920 & 1\,003\,197 \\
        Faroese & N, V, A & 3\,077 & 45\,474 \\
        Latin & N, V, A & 17\,214 & 509\,182 \\
        Hungarian & N, V, A & 13\,989 & 490\,394 \\
        Czech & N, V, A & 5\,125 & 134\,527 \\
        Lithuanian & N, V, A & 1\,458 & 136\,998 \\
        Slovenian & N, V, A & 2\,535 & 60\,110 \\
        Welsh & V & 183 & 10\,641 \\
        North Sami & N, V, A & 2\,103 & 62\,677 \\
        \hline
    \end{tabular}
    \caption{Information about the UM data for the LR languages used to train the inflectors. Livvi is composed of several dialects; we mixed them in one file. N, V and A stand for noun, verb and adjective, respectively; PoS stands for part-of-speech.}
    \label{tab:um_data}
\end{table}

\section{UD treebanks information}\label{app:tb_information}
\begin{table}[hbtp!]
    \scriptsize
     \centering
     \begin{tabular}{lrr}
     \hline
        Treebank  & \# Training sentences & \# Test sentences \\
        \hline
        Galician\textsubscript{TreeGal}  & 600 & 400\\
        Livvi\textsubscript{KKPP} & 19 & 106\\
        Faroese\textsubscript{FarPaHC} & 1\,020 & 301\\
        Latin\textsubscript{Perseus} & 1\,334 & 939\\
        Hungarian\textsubscript{Szeged} & 910 & 449\\
        Czech\textsubscript{CLTT} & 860 & 136\\
        Lithuanian\textsubscript{HSE} & 153 & 55\\
        Slovenian\textsubscript{SST} & 2\,078 & 1110\\
        Welsh\textsubscript{CCG} & 704 & 953\\
        North Sami\textsubscript{Giella} & 2\,257 & 865\\
     \hline
     \end{tabular}
     \caption{Number of training and test sentences of LR treebanks used in both experimental setups.}
     \label{tab:lr_tb}
 \end{table}

\begin{table}[hbtp!]
 \centering
 \scriptsize
 \begin{tabular}{lrr}
 \hline
    Treebank  & \# Training sentences & Post-ed.\\
    \hline
    Finnish\textsubscript{TDT} & 12\,217 & Yes \\
    Estonian\textsubscript{EDT} & 24\,633 & No \\
    Norwegian\textsubscript{Bokmaal} & 15\,696 & Yes \\
    Norwegian\textsubscript{Nynorsk} & 14\,174 & Yes\\
    Icelandic\textsubscript{IcePaHC} & 34\,007 & No \\
    Danish\textsubscript{DDT} & 4\,383 & Yes \\
    Swedish\textsubscript{LinES} & 3\,176 & Yes\\
    Polish\textsubscript{LFG} & 13\,884 & Yes\\
    Slovak\textsubscript{SNK} & 8\,483 & Yes\\
    French\textsubscript{GSD} & 14\,449 & Yes\\
    Italian\textsubscript{ISDT} & 13\,121 & Yes\\
    Romanian\textsubscript{Nonstandard} & 24\,121 & Yes\\
    Catalan\textsubscript{AnCora} & 13\,304 & Yes\\ 
    Spanish\textsubscript{AnCora} & 14\,305 & Yes\\
    Portuguese\textsubscript{Bosque} & 8\,328 & Yes\\
    Latvian\textsubscript{LVTB} & 10\,156 & Yes\\
    Bulgarian\textsubscript{BTB} & 8\,907 & Yes\\
    Croatian\textsubscript{SET} & 6\,914 & No\\
    Serbian\textsubscript{SET} & 3\,328 & No\\
    Irish\textsubscript{IDT} & 4\,005 & Yes\\
    Scottish Gaelic\textsubscript{ARCOSG} & 1\,990 & No\\
 \hline
 \end{tabular}
 \caption{Number of training sentences of the RR treebanks used in both experimental setups, and availability of post-editing for the feature conversion between UD and UM. This post-editing is introduced by \citet{mccarthy2018udw} to improve the automatic mapping of features between both schemata, as there are some discrepancies in annotation between them that makes this conversion not trivial. This process does not assure that the conversion is perfect or even good, but it improves its accuracy.}
 \label{tab:rr_tb}
\end{table}

\section{Extended results for Experiment 1}\label{app:exp1_results}
\begin{table}[hbtp!]
    \centering
    \scriptsize
\begin{tabular}{llcrr|rr}
\hline
    \multirow{2}{*}{Target treebank}      & \multirow{2}{*}{Source treebank}   & \multirow{2}{*}{ X-inflected}  &  \multicolumn{2}{c|}{GB} & \multicolumn{2}{c}{SL} \\
      &            & &   UAS &   LAS &   UAS &  LAS \\
\hline
       Czech\textsubscript{CLTT} &             Polish\textsubscript{LFG} &          No &      25.35 &      12.99 &   25.30 &   14.66 \\
       Czech\textsubscript{CLTT} &             Polish\textsubscript{LFG} &    Yes &      28.03 &      15.97 &   18.37 &    4.60 \\
       Czech\textsubscript{CLTT} &             Slovak\textsubscript{SNK} &          No &      54.59 &      44.68 &   41.51 &   30.92 \\
       Czech\textsubscript{CLTT} &             Slovak\textsubscript{SNK} &    Yes &      45.09 &      37.02 &   25.68 &   17.54 \\ \hline
  Faroese\textsubscript{FarPaHC} &             Danish\textsubscript{DDT} &          No &      30.80 &      16.14 &   23.32 &   11.80 \\
  Faroese\textsubscript{FarPaHC} &             Danish\textsubscript{DDT} &    Yes &      31.62 &      17.77 &   28.19 &   13.62 \\
  Faroese\textsubscript{FarPaHC} &      Icelandic\textsubscript{IcePaHC} &          No &      75.35 &      66.66 &   63.48 &   53.48 \\
  Faroese\textsubscript{FarPaHC} &      Icelandic\textsubscript{IcePaHC} &    Yes &      72.98 &      63.76 &   63.03 &   53.09 \\
  Faroese\textsubscript{FarPaHC} &      Norwegian\textsubscript{Bokmaal} &          No &      13.29 &       7.76 &   23.63 &   11.51 \\
  Faroese\textsubscript{FarPaHC} &      Norwegian\textsubscript{Bokmaal} &    Yes &      24.62 &      13.51 &   25.94 &   14.11 \\
  Faroese\textsubscript{FarPaHC} &      Norwegian\textsubscript{Nynorsk} &          No &      27.33 &      16.72 &   30.98 &   16.18 \\
  Faroese\textsubscript{FarPaHC} &      Norwegian\textsubscript{Nynorsk} &    Yes &      26.62 &      16.14 &   30.69 &   17.32 \\
  Faroese\textsubscript{FarPaHC} &          Swedish\textsubscript{LinES} &          No &      29.91 &      12.26 &   25.05 &    6.44 \\
  Faroese\textsubscript{FarPaHC} &          Swedish\textsubscript{LinES} &    Yes &      33.25 &      14.23 &   25.57 &    7.27 \\ \hline
 Galician\textsubscript{TreeGal} &         Catalan\textsubscript{AnCora} &          No &      54.28 &      32.96 &   42.72 &   25.66 \\
 Galician\textsubscript{TreeGal} &         Catalan\textsubscript{AnCora} &    Yes &      51.28 &      30.19 &   38.50 &   21.91 \\
 Galician\textsubscript{TreeGal} &      Portuguese\textsubscript{Bosque} &          No &      68.38 &      57.09 &   57.77 &   48.28 \\
 Galician\textsubscript{TreeGal} &      Portuguese\textsubscript{Bosque} &    Yes &      67.94 &      56.97 &   57.21 &   47.49 \\
 Galician\textsubscript{TreeGal} &         Spanish\textsubscript{AnCora} &          No &      69.37 &      49.25 &   53.51 &   36.55 \\
 Galician\textsubscript{TreeGal} &         Spanish\textsubscript{AnCora} &    Yes &      71.16 &      50.47 &   56.98 &   40.15 \\ \hline
 Hungarian\textsubscript{Szeged} &           Estonian\textsubscript{EDT} &          No &      19.13 &       3.85 &   30.62 &   10.04 \\
 Hungarian\textsubscript{Szeged} &           Estonian\textsubscript{EDT} &    Yes &      21.08 &       3.60 &   27.85 &    9.35 \\
 Hungarian\textsubscript{Szeged} &            Finnish\textsubscript{TDT} &          No &      15.20 &       4.73 &   24.05 &    4.43 \\
 Hungarian\textsubscript{Szeged} &            Finnish\textsubscript{TDT} &    Yes &      23.29 &       6.77 &   36.24 &    8.86 \\ \hline
    Latin\textsubscript{Perseus} &         Catalan\textsubscript{AnCora} &          No &      17.96 &       8.00 &   15.89 &    6.18 \\
    Latin\textsubscript{Perseus} &         Catalan\textsubscript{AnCora} &    Yes &      15.49 &       7.46 &   24.05 &   10.64 \\
    Latin\textsubscript{Perseus} &             French\textsubscript{GSD} &          No &      20.86 &       8.78 &   25.45 &    8.31 \\
    Latin\textsubscript{Perseus} &             French\textsubscript{GSD} &    Yes &      20.33 &       8.91 &   19.60 &    8.33 \\
    Latin\textsubscript{Perseus} &           Italian\textsubscript{ISDT} &          No &      21.30 &      10.10 &   19.37 &    9.19 \\
    Latin\textsubscript{Perseus} &           Italian\textsubscript{ISDT} &    Yes &      28.75 &      14.62 &   20.21 &    8.16 \\
    Latin\textsubscript{Perseus} &   Romanian\textsubscript{Nonstandard} &          No &      20.34 &       8.44 &   20.09 &    8.36 \\
    Latin\textsubscript{Perseus} &   Romanian\textsubscript{Nonstandard} &    Yes &      30.45 &      15.58 &   30.12 &   11.74 \\
    Latin\textsubscript{Perseus} &         Spanish\textsubscript{AnCora} &          No &      17.14 &       7.99 &   18.40 &    8.08 \\
    Latin\textsubscript{Perseus} &         Spanish\textsubscript{AnCora} &    Yes &      17.50 &       8.67 &   24.40 &    9.19 \\ \hline
   Lithuanian\textsubscript{HSE} &           Latvian\textsubscript{LVTB} &          No &      33.30 &      14.72 &   30.94 &    7.64 \\
   Lithuanian\textsubscript{HSE} &           Latvian\textsubscript{LVTB} &    Yes &      42.74 &      24.62 &   34.53 &   17.08 \\ \hline
       Livvi\textsubscript{KKPP} &           Estonian\textsubscript{EDT} &          No &      36.09 &      18.95 &   36.90 &   17.74 \\
       Livvi\textsubscript{KKPP} &           Estonian\textsubscript{EDT} &    Yes &      41.53 &      22.11 &   38.37 &   22.11 \\
       Livvi\textsubscript{KKPP} &            Finnish\textsubscript{TDT} &          No &      56.72 &      38.71 &   33.47 &   18.62 \\
       Livvi\textsubscript{KKPP} &            Finnish\textsubscript{TDT} &    Yes &      60.62 &      45.03 &   46.57 &   30.58 \\ \hline
North Sami\textsubscript{Giella} &           Estonian\textsubscript{EDT} &          No &      26.23 &      10.02 &   27.94 &   10.42 \\
North Sami\textsubscript{Giella} &           Estonian\textsubscript{EDT} &    Yes &      28.66 &      10.99 &   26.18 &   10.36 \\
North Sami\textsubscript{Giella} &            Finnish\textsubscript{TDT} &          No &      21.60 &       8.21 &   19.91 &    5.43 \\
North Sami\textsubscript{Giella} &            Finnish\textsubscript{TDT} &    Yes &      23.09 &       9.51 &   17.47 &    5.30 \\ \hline
    Slovenian\textsubscript{SST} &    Bulgarian\textsubscript{BTB} &          No &      38.28 &      18.96 &   30.67 &   13.09 \\
    Slovenian\textsubscript{SST} &    Bulgarian\textsubscript{BTB} &    Yes &      31.02 &      15.09 &   23.92 &   10.97 \\
    Slovenian\textsubscript{SST} &           Croatian\textsubscript{SET} &          No &      48.71 &      28.86 &   38.87 &   20.06 \\
    Slovenian\textsubscript{SST} &           Croatian\textsubscript{SET} &    Yes &      39.55 &      21.63 &   27.12 &   14.33 \\
    Slovenian\textsubscript{SST} &            Serbian\textsubscript{SET} &          No &      47.64 &      27.33 &   37.39 &   18.19 \\
    Slovenian\textsubscript{SST} &            Serbian\textsubscript{SET} &    Yes &      36.44 &      21.22 &   23.07 &   11.92 \\ \hline
        Welsh\textsubscript{CCG} &              Irish\textsubscript{IDT} &          No &      38.17 &      13.80 &   34.87 &   13.50 \\
        Welsh\textsubscript{CCG} &              Irish\textsubscript{IDT} &    Yes &      34.20 &      13.25 &   33.27 &   12.09 \\
        Welsh\textsubscript{CCG} & S. Gaelic\textsubscript{ARCOSG} &          No &      31.29 &      10.34 &   39.48 &   13.73 \\
        Welsh\textsubscript{CCG} & S. Gaelic\textsubscript{ARCOSG} &    Yes &      35.91 &      12.82 &   33.48 &   10.68 \\
\hline
\end{tabular}
    \caption{Scores obtained by the models trained on the x-inflected and original source treebank on the test sets of the target LR treebanks, for both parsers used in Experiment 1.}
    \label{tab:exp1_full_results}
\end{table}

\clearpage

\section{Extended results for Experiment 2}\label{app:exp2_results}

\begin{figure}[hbtp!]
    \centering
    \includegraphics[width=0.48\textwidth]{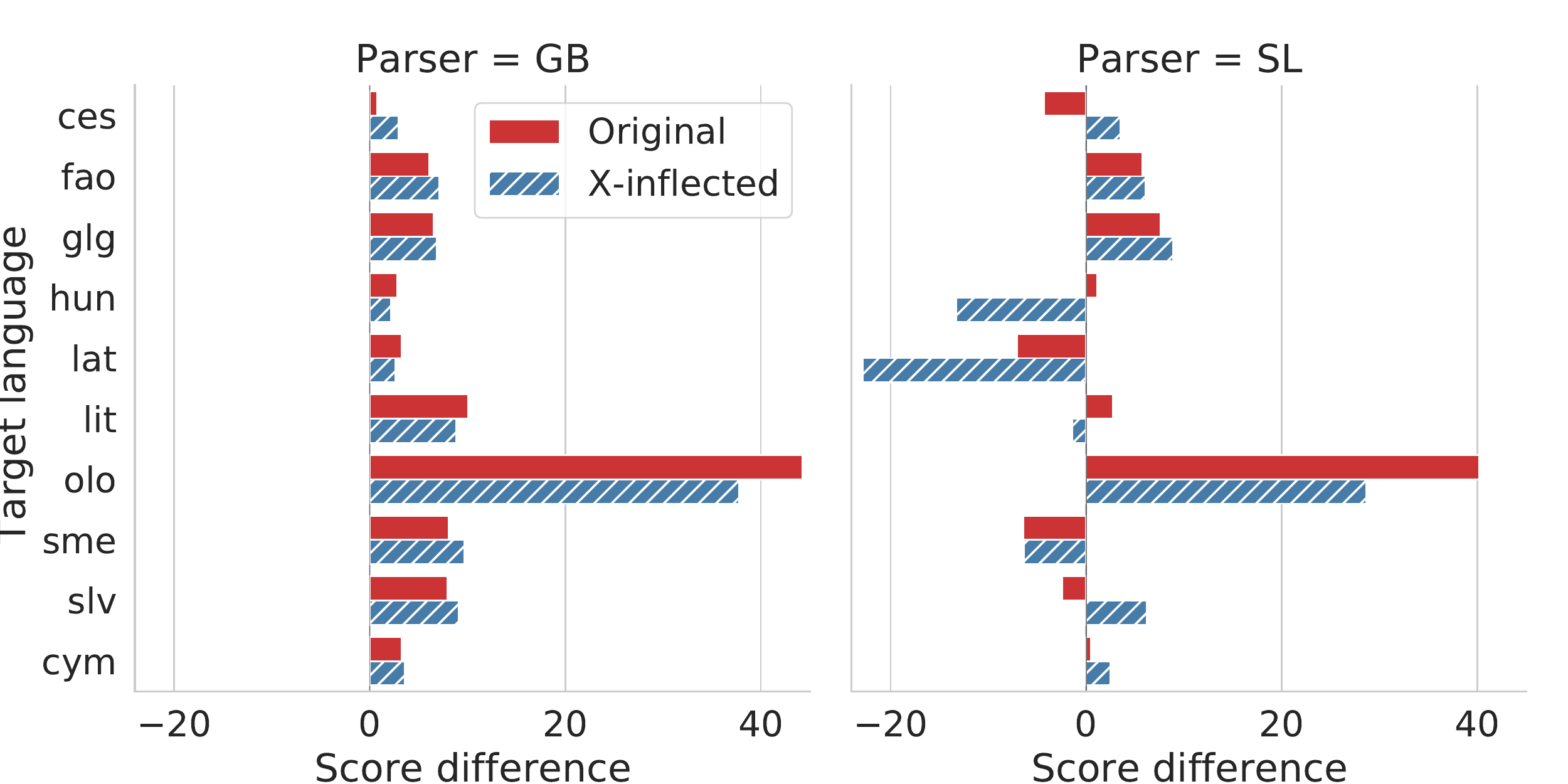}
    \caption{UAS difference between the models trained on the original and x-inflected groups \emph{with respect to} the model trained on the LR treebank, for both parsers.}
    \label{fig:fewshot}
\end{figure}

\begin{table}[hbtp]
    \centering
    \scriptsize
\begin{tabular}{lcrrrr}
\hline
     \multirow{2}{*}{Target treebank} & \multirow{2}{*}{Model} & \multicolumn{2}{c}{GB} & \multicolumn{2}{c}{SL} \\
      &       &  UAS &  LAS &  UAS &  LAS \\
\hline
       Czech\textsubscript{CLTT} & X-inflected  &      88.18 &      84.06 &   76.57 &   72.30\\
       Czech\textsubscript{CLTT} & LR Treebank  &      85.22 &      80.16 &   73.10 &   67.70\\
       Czech\textsubscript{CLTT} &    Original  &      85.99 &      82.05 &   68.77 &   63.56\\ \hline
  Faroese\textsubscript{FarPaHC} & X-inflected  &      89.95 &      87.04 &   83.35 &   79.85\\
  Faroese\textsubscript{FarPaHC} & LR Treebank  &      82.86 &      77.98 &   77.29 &   71.78\\
  Faroese\textsubscript{FarPaHC} &    Original  &      88.96 &      86.35 &   83.02 &   78.78\\ \hline
 Galician\textsubscript{TreeGal} & X-inflected  &      85.54 &      79.17 &   80.91 &   73.63\\
 Galician\textsubscript{TreeGal} & LR Treebank  &      78.69 &      71.77 &   72.02 &   63.62\\
 Galician\textsubscript{TreeGal} &    Original  &      85.22 &      79.02 &   79.61 &   72.87\\ \hline
 Hungarian\textsubscript{Szeged} & X-inflected  &      81.85 &      73.77 &   51.99 &   39.61\\
 Hungarian\textsubscript{Szeged} & LR Treebank  &      79.65 &      72.76 &   65.31 &   56.04\\
 Hungarian\textsubscript{Szeged} &    Original  &      82.47 &      74.34 &   66.47 &   53.17\\ \hline
    Latin\textsubscript{Perseus} & X-inflected  &      62.51 &      47.82 &   29.28 &   16.68\\
    Latin\textsubscript{Perseus} & LR Treebank  &      59.91 &      47.30 &   52.15 &   37.47\\
    Latin\textsubscript{Perseus} &    Original  &      63.17 &      48.76 &   45.08 &   30.63\\ \hline
   Lithuanian\textsubscript{HSE} & X-inflected  &      55.28 &      41.51 &   36.89 &   19.06\\
   Lithuanian\textsubscript{HSE} & LR Treebank  &      46.42 &      32.17 &   38.30 &   23.58\\
   Lithuanian\textsubscript{HSE} &    Original  &      56.51 &      44.53 &   41.04 &   23.87\\ \hline
       Livvi\textsubscript{KKPP} & X-inflected  &      59.74 &      43.41 &   41.13 &   25.40\\
       Livvi\textsubscript{KKPP} & LR Treebank  &      21.98 &       8.94 &   12.50 &    4.10\\
       Livvi\textsubscript{KKPP} &    Original  &      66.26 &      53.70 &   52.69 &   38.24\\ \hline
North Sami\textsubscript{Giella} & X-inflected  &      76.03 &      68.99 &   51.88 &   40.08\\
North Sami\textsubscript{Giella} & LR Treebank  &      66.32 &      58.57 &   58.21 &   46.56\\
North Sami\textsubscript{Giella} &    Original  &      74.42 &      66.94 &   51.77 &   40.85\\ \hline
    Slovenian\textsubscript{SST} & X-inflected  &      73.95 &      67.09 &   66.88 &   58.21\\
    Slovenian\textsubscript{SST} & LR Treebank  &      64.86 &      55.98 &   60.70 &   50.88\\
    Slovenian\textsubscript{SST} &    Original  &      72.79 &      65.88 &   58.26 &   47.25\\ \hline
        Welsh\textsubscript{CCG} & X-inflected  &      86.28 &      78.06 &   81.08 &   69.76\\
        Welsh\textsubscript{CCG} & LR Treebank  &      82.71 &      72.76 &   78.60 &   66.80\\
        Welsh\textsubscript{CCG} &    Original  &      85.95 &      77.65 &   79.09 &   62.00\\
\hline
\end{tabular}
    \caption{Scores obtained by both parsers for Experiment 2 and trained on the x-inflected merged treebanks, on the original merged treebanks (baseline I), and on the LR target treebank (baseline II). All the models are evaluated on the test sets of the target LR treebanks.}
    \label{tab:my_label}
\end{table}

\section{Hardware used and computational costs}\label{app:hardware}
We used 2 GeForce RTX 3090 to train both the parsing and the morphological inflection models. For Experiment 1, we trained 56 RR parsing models. For Experiment 2, we trained 20 merged models and 10 LR parsing models. For inflection, we trained 10 morphological inflection models. For the GB parser, models took around 2-3 hours for the LR models, 4-6 for the RR models, and around 8-12 for the biggest merged models. For the SL parser, models took around 1-2 hours for the LR models, 3-4 for the RR models, and around 6-8 for the biggest merged models. In the case of inflectors, training times varied between 1 and 12 hours approximately.
 \end{document}